\documentclass[conference]{IEEEtran}
\usepackage{times}

\usepackage[numbers]{natbib}
\usepackage{multicol}
\usepackage[bookmarks=true]{hyperref}
\usepackage{graphicx}
\usepackage{amsmath}
\usepackage{amssymb}
\usepackage{color}

\begin{document}
	
\title{DA-RNN: Semantic Mapping with Data Associated Recurrent Neural Networks}




%
\author{\authorblockN{Yu Xiang and Dieter Fox}
\authorblockA{Paul G. Allen School of Computer Science \& Engineering \\
University of Washington \\
\{yuxiang, fox\}@cs.washington.edu}}

\maketitle

\begin{abstract}
3D scene understanding is important for robots to interact with the 3D world in a meaningful way. Most previous works on 3D scene understanding focus on recognizing geometrical or semantic properties of a scene independently. In this work, we introduce Data Associated Recurrent Neural Networks (DA-RNNs), a novel framework for joint 3D scene mapping and semantic labeling. DA-RNNs use a new recurrent neural network architecture for semantic labeling on RGB-D videos. The output of the network is integrated with mapping techniques such as KinectFusion in order to inject semantic information into the reconstructed 3D scene. Experiments conducted on real world and synthetic RGB-D videos demonstrate the superior performance of our method.\end{abstract}

\IEEEpeerreviewmaketitle

\section{Introduction}

For many tasks, robots need to understand the 3D structure and semantics of their environment. For example, recognizing the free space and surfaces in a scene helps motion planning in robot navigation and manipulation tasks. Semantic understanding, beyond pure geometry, enables a robot to reason about objects, which is particularly important for manipulation and human robot interaction tasks. Over the last years, various techniques have been proposed for dense 3D scene reconstruction using depth cameras, including RGBD-Mapping, KinectFusion, Kintinuous, and ElasticFusion \cite{henry2012rgb,newcombe2011kinectfusion,Whe12Kin,whelan2015elasticfusion}.  These methods jointly reconstruct the 3D scene and track the camera position from RGB-D videos. However, they do not provide semantic information about the scene. In parallel, different approaches for recognizing scene semantics have been proposed. These include methods in object detection \cite{felzenszwalb2008discriminatively,girshick2014rich}, object pose estimation \cite{brachmann2014learning,savarese20073d,xiang2012estimating}, and semantic labeling \cite{ren2012rgb,long2015fully}. Most of these methods focus on detecting specific objects or on recognizing scene elements in individual 2D images.

The goal of our work is to use RGB-D videos to reconstruct and label every observed surface element in a 3D scene, providing dense information about small objects, such as bowls and mugs, and larger objects such as tables and chairs. In such a setting, the key question is how the information from the RGB-D frames can be combined to improve recognition accuracy. Recent approaches handle this by incorporating recognition results from individual RGB-D frames into a 3D model, possibly followed by additional reasoning over the 3D structure~\cite{Lai13Obj,lai2014unsupervised,mccormac2016semanticfusion}. However, in these approaches, the reasoning about individual frames and their information accumulation is only loosely coupled.

In this work, we introduce DA-RNNs, a deep network architecture that tightly connects the analysis of individual RGB-D frames and their integration over time.  To do so, we take advantage of \emph{Recurrent} Neural Networks (RNNs), where recurrent units connect information over time.  A naive approach for achieving a strong connection between the mapping and the labeling process would be to establish a fixed network structure in 3D and treat each surface element in a KinectFusion or ElasticFusion map as a recurrent unit in the RNN. Unfortunately, such an approach is not feasible since it would quickly exhaust the memory available even on large-scale GPUs. To overcome this problem, our approach performs recurrent reasoning only over those parts of the map that are currently observed by the RGB-D camera.  Specifically, we introduce a new recurrent unit inside our RNN called Data Associated Recurrent Unit (DA-RU). Each DA-RU corresponds to a pixel in the input image. The hidden state of the DA-RU accumulates information about that pixel in time.  Crucially, the temporal connectivity between the DA-RU states of consecutive frames is not fixed, but depends on the data association provided by the mapping process.  As a result, each DA-RU incorporates the hidden state from the associated DA-RU in the previous frame, allowing information to flow in a spatially consistent way.


\begin{figure}
	\centering
	\includegraphics[height = 0.55\linewidth, width = \linewidth]{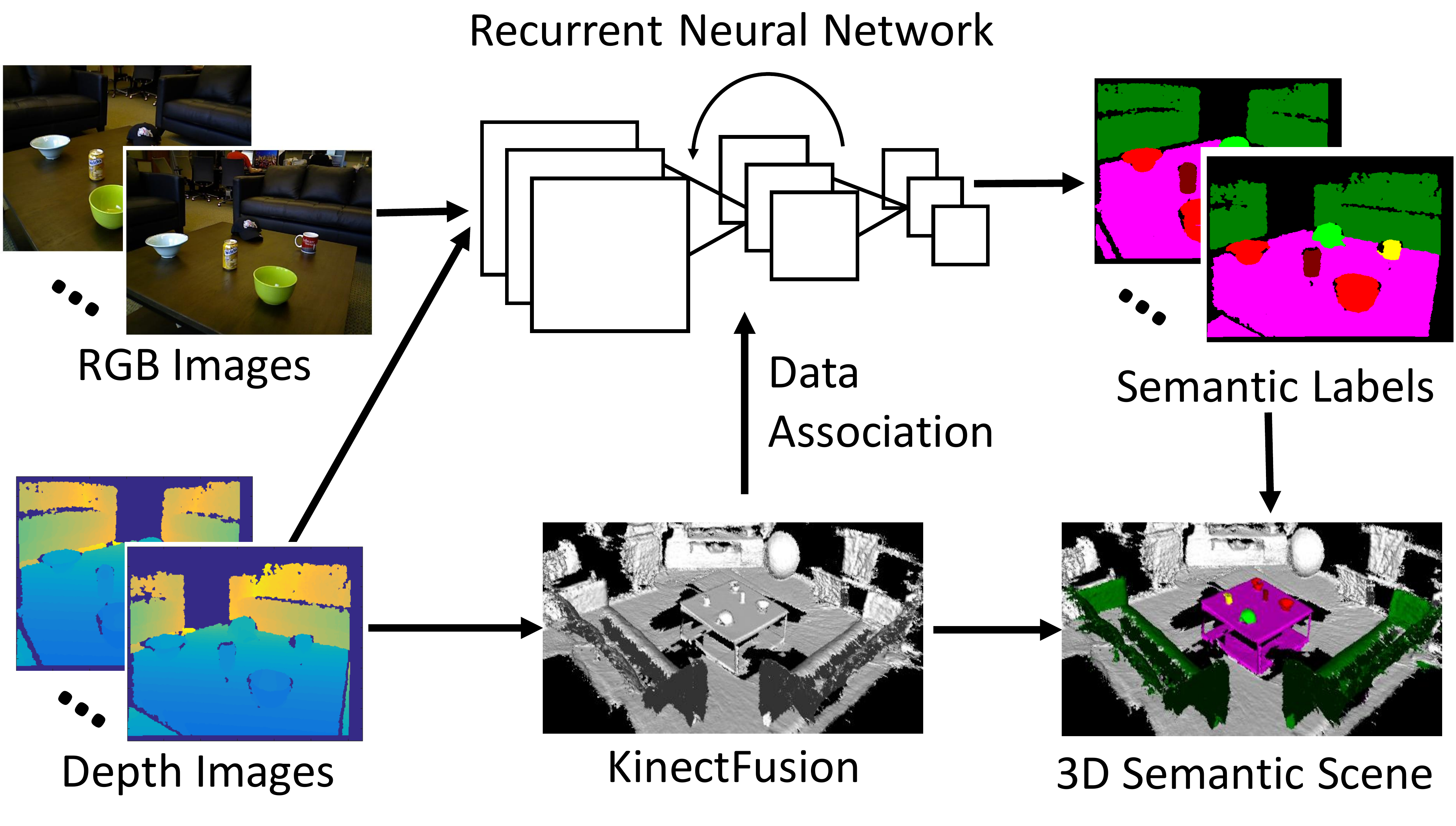}
	\caption{Overview of the DA-RNN framework.  RGB-D frames are fed into a Recurrent Neural Network.  KinectFusion provides the 3D reconstruction and  the data associations necessary to connect recurrent units between RGB-D frames.  The pixel labels provided by the RNN are integrated into the 3D semantic map. The overall labeling and reconstruction process runs at 5fps.}
	\label{fig:framework}
	\vspace{-4mm}
\end{figure}

In order to semantically reconstruct the 3D scene, we integrate the outputs of our DA-RNN into the 3D voxels of KinectFusion, which provides a consistent semantic labeling of the 3D scene (similar to~\cite{lai2014unsupervised,mccormac2016semanticfusion}). Fig.~\ref{fig:framework} illustrates an overview of our framework.  

We have conducted extensive experiments to test our framework on the RGB-D Scene dataset \cite{lai2014unsupervised} and a synthetic dataset we generated with 3D shapes from the ShapeNet repository \cite{chang2015shapenet}. The experimental results demonstrate that DA-RNNs are able to provide superior semantically labeled 3D scenes from RGB-D videos. Our code and data are available at \url{https://rse-lab.cs.washington.edu/projects/darnn/}.

In summary, our work has the following key contributions:
\begin{itemize}	
	\item We propose a novel recurrent neural network for semantic labeling on RGB-D videos with a new data associated recurrent unit to capture dependencies across video frames.
	\item We introduce a novel updating rule for DA-RU's to perform weighted moving averaging of the hidden state. 
	\item We integrate DA-RNN's with KinectFusion for semantic 3D scene reconstruction.
	\item We contribute pixel-wise semantic labels on the RGB-D Scene dataset \cite{lai2014unsupervised} and a new synthetic dataset which can benefit future research on 3D semantic mapping.
\end{itemize} 

This paper is organized as follows. After discussing related work, we introduce DA-RNNs, followed by experimental results and a conclusion.

\section{Related Work}

Our work is mostly related to 3D mapping and semantic labeling methods in the literature.

\subsection{Dense 3D Scene Reconstruction}

3D reconstruction techniques can be roughly classified into point-based methods, voxel-based methods and surfel-based methods. Point-based methods use 3D points to represent  3D scenes~\cite{snavely2008skeletal,crandall2011discrete,henry2012rgb}. Voxel-based methods such as KinectFusion, PatchVolumes, or Kintinuous \cite{newcombe2011kinectfusion,henry2013patch,Whe12Kin} employ a volumetric representation of the 3D space, which reconstruct dense 3D surfaces of the scene. Surfel-based methods \cite{keller2013real,henry2012rgb,whelan2015elasticfusion} make a trade-off between 3D points and voxels, where the 3D scene is represented compactly by 3D disks, i.e., surfels.

In principle, our DA-RNN framework only requires dense data associations between consecutive frames.  It is thus independent of the underlying representation and could be combined with any of the reconstruction techniques described above. Here, we use KinectFusion~\cite{newcombe2011kinectfusion} to achieve a volumetric representation for geometry and semantics. 

\subsection{Semantic Labeling}

Semantic labeling on images classifies each pixel of an input image into one of the predefined semantic classes. The semantic labeling problem has often been tackled with probabilistic graphical models such as Markov Random Fields (MRFs) or Conditional Random Fields (CRFs) \cite{shotton2006textonboost,krahenbuhl2011efficient}, which model the context around pixels. More recently, convolutional neural networks have been applied to semantic labeling \cite{long2015fully,zheng2015conditional,badrinarayanan2015segnet,chen2016deeplab}, which achieve significant improvement over previous methods. However, all these approaches mainly focus on semantic labeling of a single image. Recurrent neural networks \cite{pavel2015recurrent,shelhamer2016clockwork} have been applied to semantic video segmentation, which exploit the temporal relationship or information provided by multiple viewpoints of a scene. \cite{Lai13Obj,mccormac2016semanticfusion} show how the labels extracted from individual RGB-D frames can be incorporated into a voxel or surfel map, resulting in more stable labeling. Further improvements are achieved by performing MRF or CRF inference in the 3D map. Approaches such as \cite{salas2013slam++,lai2014unsupervised,song2016deep} perform labeling by conducting 3D object detection through the 3D reconstruction, thereby potentially incorporating information that is not available in any single view.

Different from these works, we propose a recurrent neural network architecture that tightly integrates the information contained in multiple viewpoints of an RGB-D video stream.  Both individual frame and across frame parameters are learned in a single network structure. In contrast to existing RNNs, DA-RNNs do not assume a fixed relationship between input images and network structure, but rely on data association to generate the connections between recurrent units on the fly.  The recurrent layer we introduce in this work could also be used as a standalone layer and plugged into existing CNN-based methods for semantic video labeling.

\section{Method}

\begin{figure*}
	\centering
	\includegraphics[height = 0.22\linewidth, width = \linewidth]{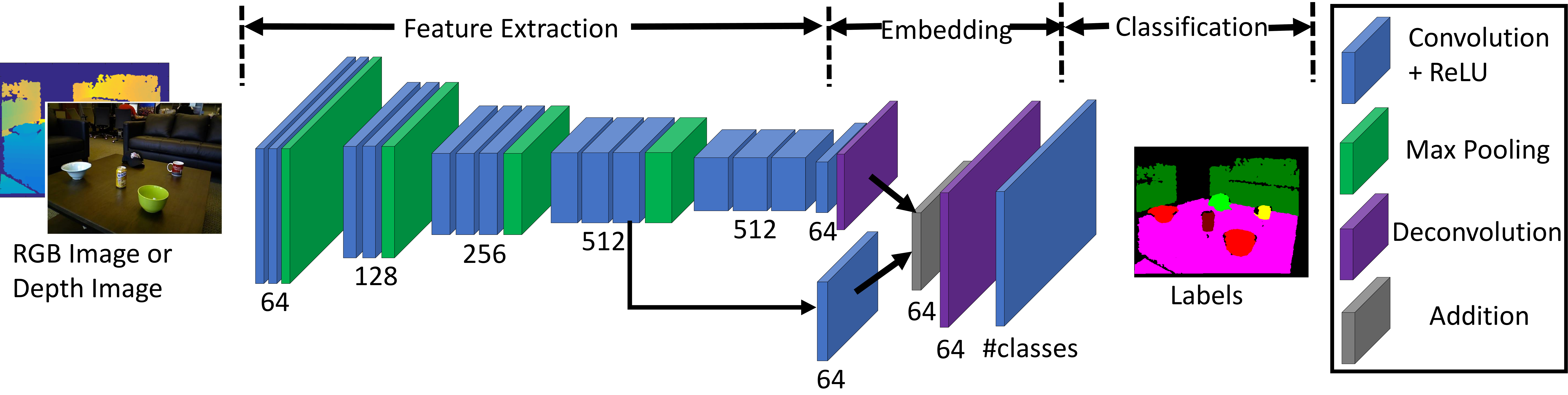}
	\caption{Architecture of our single stream network for semantic labeling.}
	\label{fig:single_stream}
	\vspace{-3mm}
\end{figure*}

\begin{figure*}
	\centering
	\includegraphics[height = 0.28\linewidth, width = \linewidth]{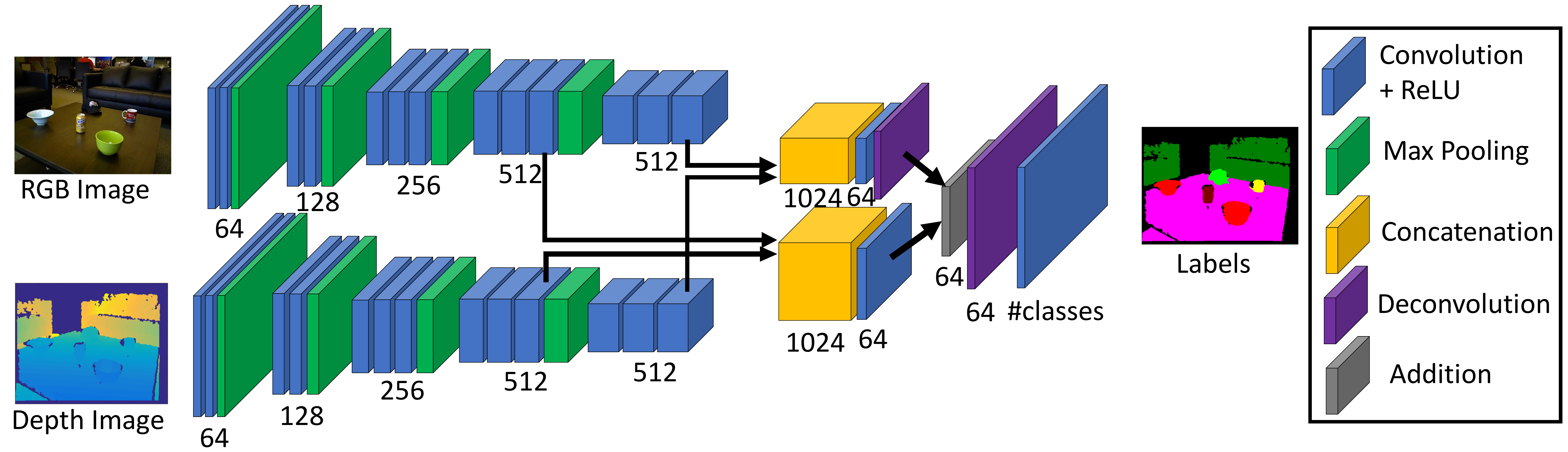}
	\caption{Architecture of our double stream network for semantic labeling.}
	\label{fig:double_stream}
	\vspace{-4mm}
\end{figure*}

In this section, we present our framework for 3D semantic mapping using RGB-D videos. We first describe our design of the convolutional neural network for single frame semantic labeling. Then, we extend the single frame network to a recurrent neural network for semantic labeling on videos. Finally, we integrate the recurrent neural network with KinectFusion \cite{newcombe2011kinectfusion} in order to semantically reconstruct the 3D scene.

\subsection{Single Frame Labeling with Fully Convolutional Networks}

The basis of our semantic labeling framework is a Fully Convolutional Network (FCN) for single frame labeling. An influential network architecture for semantic labeling as been introduced by \cite{long2015fully}, which converts a network for image classification into fully convolutional by treating the fully connected layers in the network as $1 \times 1$ convolutional layers. In addition, \cite{long2015fully} uses deconvolutional layers to increase the resolution of the network output. Inspired by \cite{long2015fully}, we design our network architecture for single frame labeling as illustrated in Fig. \ref{fig:single_stream}.

\subsubsection{Single Stream Network}

Our single stream network in Fig. \ref{fig:single_stream} takes a single tensor as input, such as an RGB image or a depth image. It consists of 16 convolutional layers, 4 max pooling layers, 2 deconvolutional layers and 1 addition layer. All the convolutional filters in the network are of size $3 \times 3$ and stride $1$. The max pooling layers are of size $2 \times 2$ and stride $2$. Therefore, each max pooling layer reduces the resolution of its input by a factor of 2. The output of the 4th max pooling layer is 16 times smaller than the input image. The first deconvolutional layer doubles the resolution of its input, while the second deconvolutional layer increases the resolution by 8 times. As a result, the output of the network has the same resolution as the input image, i.e., dense pixel-wise labeling.

We design the network architecture with three phases as in Fig. \ref{fig:single_stream}. The first 13 convolutional layers and the 4 max pooling layers are considered to be the feature extraction phase, which extracts 512-dimensional feature vectors for the input image. The second phase is the embedding phase, which embeds the 512-dimensional features into a 64-dimensional space while increasing the resolution of the feature map using deconvolutional layers. A skip link is used in the embedding phase to combine features from an earlier convolutional layer motivated by \cite{long2015fully} (i.e., the one before the 4th max pooling layer). The last phase of the network classifies each pixel into a semantic class using a convolutional layer. The output of this convolutional layer is treated as the labeling scores for pixels, which has $n$ channels with $n$ the number of the semantic classes. By applying a softmax layer on the labeling scores, we can obtain the class probabilities of the pixels.

\subsubsection{Double Stream Network}

When the input data is multimodal such as color and depth, we have designed the double stream network to fuse RGB-D data (Fig. \ref{fig:double_stream}). In this network, the RGB image and the depth image are processed separately with different convolutional layers for feature extraction. These layers share the same structure as the feature extraction phase in the single stream network. To combine the two types of features, we introduce two concatenation layers, which stack the 512-dimensional features from the RGB image and the depth image and generate 1024-dimensional features. These features are embedded into a 64-dimensional space and classified as in the single stream network. Note that we utilize the ``late fusion'' strategy in this network, where features for color and depth are computed independently and then concatenated.

\subsection{Video Semantic Labeling with DA-RNNs}

In videos, due to the smooth change in camera motion or object motion, information flows across video frames. How to effectively utilize the temporal information for semantic labeling in videos is still an open question. In this work, we propose a Data Associated Recurrent Neural Network (DA-RNN) for video semantic labeling which stores and passes information across frames.

\subsubsection{DA-RNN Architecture}

\begin{figure*}
	\centering
	\includegraphics[height = 0.5\linewidth, width = \linewidth]{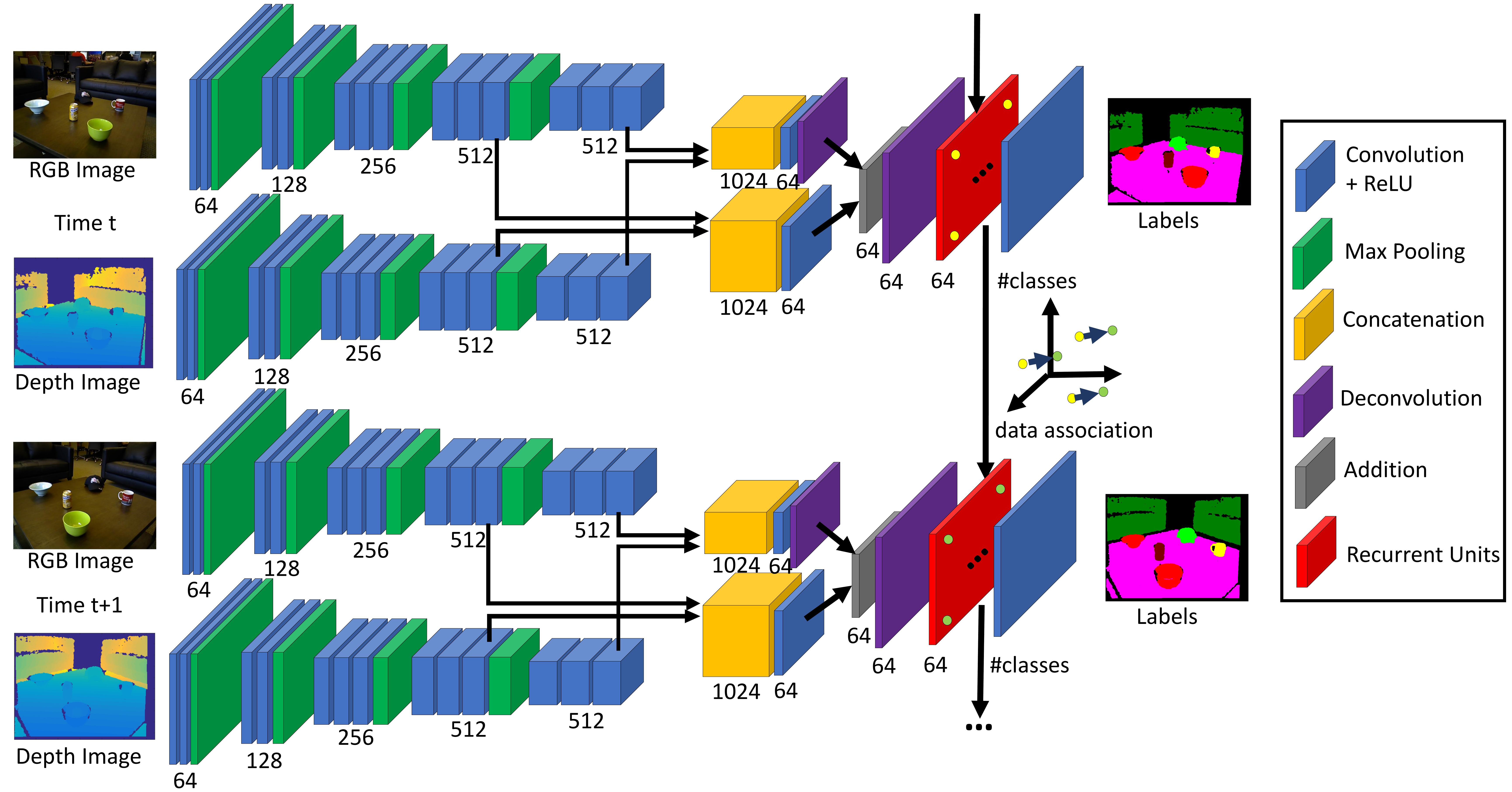}
	\caption{Architecture of our data associated recurrent neural network for semantic labeling on videos.}
	\label{fig:rnn}
	\vspace{-2mm}
\end{figure*}

The architecture of our DA-RNN for semantic labeling is illustrated in Fig.~\ref{fig:rnn}. Based on our double stream network for single frame labeling, we introduce a recurrent layer which takes the embedded features of the current frame as input and generates new features for classification. The recurrent layer is designed to combine features from the previous frames and features in the current frame in order to utilize information across frames.

Specifically, the recurrent layer contains one recurrent unit for each pixel location. These recurrent units maintain and update their hidden states, storing information from previous frames. The outputs of the recurrent units depend on their inputs and hidden states. Two widely used recurrent units are the Long Short-Term Memory (LSTM) unit \cite{hochreiter1997long} and the Gated Recurrent
Unit (GRU) \cite{cho2014properties}. Both LSTM and GRU have been shown to perform well in tasks that require capturing long-term dependencies, such as natural language processing, speech recognition and machine translation \cite{sundermeyer2012lstm,graves2013hybrid,sutskever2014sequence}. However, both of them employ the hyperbolic tangent function in updating their hidden states, which makes the gradient back-propagation training inefficient. In DA-RNN, we introduce a new recurrent unit, which is explicitly designed to fuse features across video frames and can be trained more efficiently by using the Rectified Linear Unit (ReLU) as the activation function.

\subsubsection{Data Associated Recurrent Unit (DA-RU)}

\begin{figure}
	\centering
	\includegraphics[height = 0.55\linewidth, width = \linewidth]{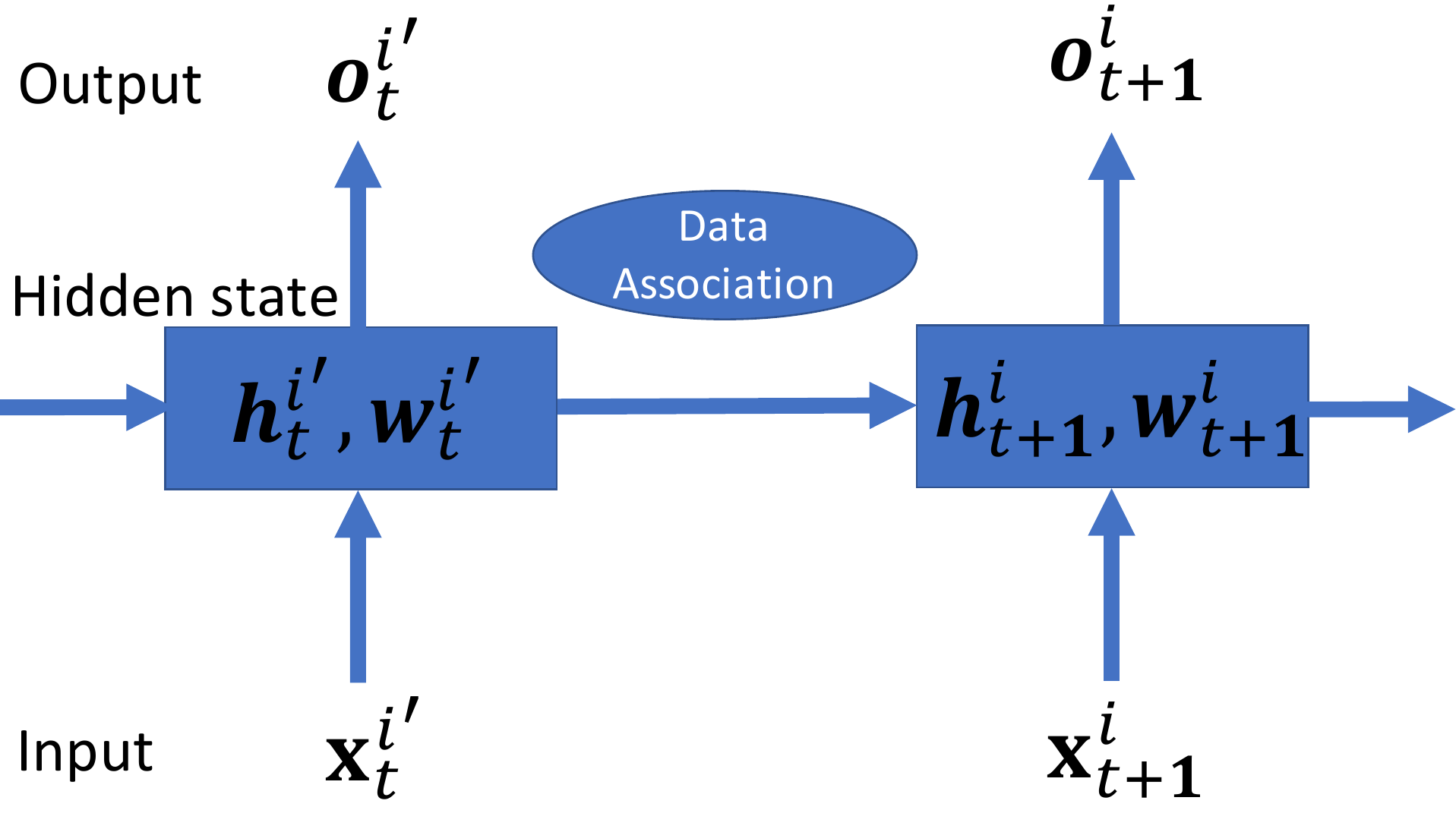}
	\caption{The block diagram of our Data Associated Recurrent Unit (DA-RU).}
	\label{fig:daru}
	\vspace{-4mm}
\end{figure}

The recurrent layer in our RNN contains $N$ recurrent units, where $N$ is the number of pixels in the input image. At time $t$, the $i$th recurrent unit stores a pair of vectors $\left< \mathbf{h}_t^i, \mathbf{w}_t^i \right>, i = 1,\ldots,N$, where $\mathbf{h}_t^i$ denotes the hidden state of the unit, and $\mathbf{w}_t^i$ indicates the weight vector for the hidden state. $\mathbf{h}_t^i$ and $\mathbf{w}_t^i$ have the same dimension (64-D in our RNN). We can interpret $\mathbf{w}_t^i$ as measuring the importance of the elements of the hidden state. At time $t + 1$, given input $\mathbf{x}_{t+1}^{i}$ from the previous layer (the second deconvolutional layer in our RNN), unit $i$ updates its hidden state and weight vector, and generates its output according to a set of rules described below. Fig. \ref{fig:daru} illustrates the block diagram of the DA-RU.

\textbf{Data association: }
\begin{equation} \label{eq:asso}
\left< \widetilde{\mathbf{h}}_{t+1}^{i}, \widetilde{\mathbf{w}}_{t+1}^{i} \right> = 
\left \{
\begin{aligned}
\left< \mathbf{0}, \mathbf{0} \right>, &\text{ if no association}\\
\left< \mathbf{h}_t^{i'}, \mathbf{w}_t^{i'} \right>, &\text{ if $p_{t+1}^i$ associated to $p_{t}^{i'}$},
\end{aligned} \right.
\end{equation}
where $p_{t+1}^i$ and $p_{t}^{i'}$ denote the corresponding pixels of unit $i$ at time $t+1$ and unit $i'$ at time $t$ respectively. Eq. \eqref{eq:asso} indicates that a unit at time $t$ passes its hidden state and weight vector to a unit at time $t+1$ via data association between pixels. If a unit at time $t+1$ is not associated with any previous unit, its hidden state and weight vector are initialized as zeros. All the units in the first frame of a video are initialized with zeros.

\textbf{Computing weights for the input: }
\begin{equation}
\widehat{\mathbf{w}}_{t+1}^{i} = \sigma (\mathbf{W} [ \widetilde{\mathbf{h}}_{t+1}^{i}, \mathbf{x}_{t+1}^{i} ] + \mathbf{b} ), \label{eq:weight}
\end{equation}
where $\widehat{\mathbf{w}}_{t+1}^{i}$ is the weight vector for the input $\mathbf{x}_{t+1}^{i}$, which is a function of the hidden state from the previous frame and the input of the current frame.  $\mathbf{W},\mathbf{b}$ are the parameters of the recurrent layer, which are shared by all the units in the layer, $\sigma(\cdot)$ indicates the logistic sigmoid function, $[\cdot, \cdot]$ denotes concatenation of two vectors. $\mathbf{W}$ is a $d \times 2d$ matrix and $\mathbf{b}$ is a $d$ dimensional bias vector, where $d$ is the dimension of the hidden state.

\textbf{Updating weight vector:}
\begin{equation} 
\mathbf{w}_{t+1}^{i} = \widetilde{\mathbf{w}}_{t+1}^{i} + \widehat{\mathbf{w}}_{t+1}^{i}.
\end{equation}
The weight vector at time $t+1$ is the sum between the accumulated weight vector from the previous frame and the weight vector for the current input.

\textbf{Updating hidden state:}
\begin{equation} \label{eq:hidden}
\mathbf{h}_{t+1}^{i} = f( (\widetilde{\mathbf{w}}_{t+1}^{i} \oslash \mathbf{w}_{t+1}^{i}) \otimes \widetilde{\mathbf{h}}_{t+1}^{i}  +  (\widehat{\mathbf{w}}_{t+1}^{i} \oslash \mathbf{w}_{t+1}^{i}) \otimes \mathbf{x}_{t+1}^{i} ),
\end{equation}
where $f(x) = \max(0, x)$ is the Rectified Linear Unit (ReLU) activation function, and $\oslash, \otimes$ denotes element-wise division and element-wise multiplication between vectors respectively. As we can see from Eq. \eqref{eq:hidden}, the new hidden state is computed as a weighted sum between the hidden state from the previous frame and the input for the current frame, where the weights are accumulated in time. 

\textbf{Computing Output:}
\begin{equation} 
\mathbf{o}_{t+1}^{i} = g(\mathbf{h}_{t+1}^{i}).
\end{equation}
The output of the unit is defined as a function $g(\cdot)$ of the hidden state. In our RNN, we simply use the hidden state as the output of the unit, i.e., $g(\cdot)$ is the identity function.

We name the aforementioned recurrent unit as the Data Associated Recurrent Unit (DA-RU). DA-RU performs weighted moving average of its input in time via data association, where the weights are dynamically generated based on the parameters of the unit and the data it receives, while the parameters are learned during network training. In DA-RNNs, the DA-RUs are used to combine features from the previous frames and features in the current frame for semantic labeling (Fig. \ref{fig:rnn}).   

\subsection{Joint 3D Mapping and Semantic Labeling}

In DA-RNNs, data association is needed in order to associate DA-RUs in the recurrent layer across video frames. In general, different data association algorithms can be applied, such as optical flow on RGB images or Iterative Closest Point (ICP) on depth images. In this work, we integrate DA-RNNs with KinectFusion \cite{newcombe2011kinectfusion}, a dense 3D mapping technique using depth camera. KinectFusion estimates the camera poses of the video frames, from which we can compute the data association for the recurrent layer in the RNN. In addition, we fuse the semantic labels of pixels into the volumetric space in KinectFusion. Consequently, our system is able to reconstruct and semantically label the 3D scene.

\subsubsection{Data Association with KinectFusion}

KinectFusion represents the 3D scene with a 3D voxel space which stores the values of the Truncated Signed Distance Function (TSDF). The TSDF value of a voxel indicates the signed distance from the voxel to the closest zero crossing, i.e., surface. Given a stream of depth images, these TSDF values are updated per-frame. In order to fuse the depth images into the voxel space, KinectFusion performs camera tracking by estimating the 6DOF camera pose for each frame. The camera pose estimation is achieved by performing ICP between the 3D points from the current depth image and the 3D points extracted from the surface of the KinectFusion map. Given the camera poses of two consecutive frames from KinectFusion, we compute the data association between the two frames by back-projecting one frame into 3D points in the KinectFusion map and then projecting these 3D points onto the other frame using the estimated camera poses.

\subsubsection{Semantic Fusion}

By combining DA-RNNs with 3D mapping techniques such as KinectFusion, we are able to propagate semantic information into the 3D space. In addition, the semantic labels from different views of the same 3D location are fused in order to obtain a consistent understanding of the 3D space. For each voxel in KinectFusion, we store a probability vector of the semantic label space in addition to the TSDF value. Given a new depth image, the TSDF values of the voxels are first updated as in the traditional KinectFusion. Then, for voxels whose signed distances are smaller than the truncated distance threshold, i.e., voxels around the surface, their probability vectors are updated using the probability map of the semantic labels predicted by the RNN. A running averaging is used for both the TSDF and the probability vector to reduce noise in the fusing process. At any time step, the label of a voxel is predicted as the semantic class with the maximum probability according to the stored probability vector in the voxel. Fig. \ref{fig:framework} illustrates the 3D mapping and semantic labeling pipeline of our framework.

\section{Experiments}

In this section, we conduct experiments to evaluate our proposed system for 3D scene mapping and semantic labeling.

\subsection{Datasets}

\subsubsection{RGB-D Scene Dataset}

Two RGB-D video datasets are used to test our method. The first one is the RGB-D Scene dataset introduced by \cite{lai2014unsupervised}, which consists of 14 RGB-D videos captured by Kinect in indoor scenes. Each scene is reconstructed as a 3D point cloud aligned via the Patch Volumes Mapping method \cite{henry2013patch}. Then these 3D point clouds are labeled by 9 object class labels plus background. However, the RGB-D Scene dataset does not provide pixel-wise labeling for every video frame, which is needed in order to train and test our RNN. We could project the labeled 3D points onto video frames, but the projection only provides sparse labeling of the frames, i.e., not every pixel is labeled. Instead, we use the following procedure to obtain dense labeling on the RGB-D Scene dataset.

Each scene is first reconstructed with KinectFusion. After the 3D reconstruction, we manually annotate the 3D bounding box of the object in the 3D map. For each depth image, we convert it into a 3D point cloud and transform the 3D point cloud into the reconstructed 3D space according to its camera pose estimated from KinectFusion. Finally, 3D points inside a 3D bounding box are labeled with the semantic class of the object inside the box. In this way, we obtain dense labels for all the depth images. Note that pixels with missing depth values are not labeled according to our labeling procedure. We use 7 videos for training (5,808 frames) and the other 7 videos for testing (5,619 frames).

\subsubsection{ShapeNet Scene Dataset}

The second dataset is a synthetic dataset we generated using 3D shapes from the ShapeNet repository \cite{chang2015shapenet}. We selected 3D shapes with high quality and texture in 7 object categories: bottle (110 objects), can (25 objects), cap (23 objects), keyboard (36 objects), monitor (95 objects), mug (65 objects) and table (508 objects). We first compose virtual scenes, each with a table on the ground and five table-top objects among bottle, can, cap, keyboard, monitor and mug. Then, we render each virtual scene into RGB images and depth images from a simulated camera trajectory around the table. To obtain the semantic labels of the rendered images, we color the 3D shapes with distinct colors for each class and render the colored scene again using the same camera trajectory. By checking the color of the pixels in this second-pass rendering, we obtain the class labels of the images.

In total, we generated 100 virtual scenes, i.e., 100 RGB-D videos, by randomly sampling 3D shapes from the 7 object categories. 100 frames are rendered for each scene from a sampled camera trajectory. We use 50 videos for training (5,000 frames) and test on the other 50 videos (5,000 frames). Different from the RGB-D Scene dataset, we make sure that there is no overlapping object instances appearing in both the training set and the test set.

\subsection{Evaluation Metrics}

We evaluate our method on semantic labeling of pixels and 3D points. For pixel labeling, we adopt the pixel Intersection over Union (IoU) as the evaluation metric, which is the standard metric used for image semantic labeling. Pixel IoU computes the intersection over union of the predicated pixel labels and the ground truth pixel labels on the entire dataset for every class. For 3D point labeling, we follow \cite{lai2014unsupervised} and use the precision and recall on 3D points as the evaluation metric in order to achieve a fair comparison.

\subsection{Implementation Details}

DA-RNN is implemented using the TensorFlow library \cite{abadi2016tensorflow} with Python interface for communication with the KinectFusion module. In training, the parameters of the first 13 convolutional layers in the feature extraction phase are initialized with the VGG16 network \cite{simonyan2014very} trained on ImageNet \cite{deng2009imagenet}. Learning is conducted by Stochastic Gradient Descent (SGD) with momentum, where the loss function is the softmax cross entropy loss for pixels. For our single stream FCN and double stream FCN, each SGD mini-batch is a single image, chosen uniformly at random. For DA-RNN, each SGD mini-batch is a  video sequence of 3 consecutive frames. In testing, video frames are processed sequentially, and the hidden states of the DA-RNN are passed to the next frame via data association for the entire video sequence. In this way, the DA-RNN captures long term dependencies between pixels.

\subsection{Comparison on Network Architectures}

\begin{table} \setlength{\tabcolsep}{2pt}
	{
		\centering{
			\begin{tabular}{|l||c|c|c|c|}
				\hline Methods  & FCN \cite{long2015fully}  & Our FCN & Our GRU-RNN & Our DA-RNN \\
				\hline
				\hline Background & 94.3 & 96.1 & 96.8  & \textbf{97.6} \\
				\hline Bowl       & 78.6 & 87.0 & 86.4 & \textbf{92.7} \\
				\hline Cap        & 61.2 & 79.0 & 82.0 & \textbf{84.4} \\
				\hline Cereal Box & 80.4 & 87.5 & 87.5 & \textbf{88.3} \\
				\hline Coffee Mug & 62.7 & 75.7 & 76.1 & \textbf{86.3} \\
				\hline Coffee Table & 93.6 & 95.2 & 96.0  & \textbf{97.3} \\
				\hline Office Chair & 67.3 & 71.6 &  72.7 & \textbf{77.0} \\
				\hline Sofa Can   & 73.5 & 82.9 &  81.9  & \textbf{88.7} \\
				\hline Sofa       & 90.8 & 92.9 &  93.5  & \textbf{95.6} \\
				\hline Table      & 84.2 & 89.8 &  90.8  & \textbf{92.8} \\
				\hline
				\hline MEAN       & 78.7 & 85.8 &  86.4  & \textbf{90.1} \\
				\hline
			\end{tabular}
			\caption{Comparison in network architectures for image pixel labeling on the RGB-D Scene dataset. The network input is RGB image.}
			\label{table:arch}
		}
		\vspace{-8mm}
	}
\end{table}

In this experiment, we fix the network input to RGB images and compare different network architectures for pixel-wise semantic labeling. Table \ref{table:arch} presents the pixel IoU on the RGB-D Scene dataset for four different networks. 

i) We compare our single stream FCN (Fig. \ref{fig:single_stream}) with the FCN in \cite{long2015fully} which is fine-tuned on the RGB-D Scene dataset using the same experimental setup as ours. As we can see from the table, our single stream FCN significantly outperforms the FCN in \cite{long2015fully}. \cite{long2015fully} converts the VGG16 network into a fully convolutional network for semantic labeling. There are five max pooling layers and two 4096-dimensional fully connected layers (eventually converted to $1 \times 1$ convolutional layers) in the network, which outputs blob-like segmentations and cannot capture fine-grained details of the objects. In contrast, our FCN uses fewer max pooling layers and embeds the convolutional features into a low dimensional space (64-D) before classification, which is able to generate shaper segmentations of the objects.

ii) We compare two types of recurrent unit in our RNN architecture: Gated Recurrent Unit (GRU) \cite{cho2014properties} and the DA-RU we introduce in this work. From Table \ref{table:arch}, we can see that our DA-RU achieves better labeling performance than GRU. First, our DA-RU can be trained more efficiently since it uses the ReLU function instead of the hyperbolic tangent function in updating its hidden state. Second, the DA-RU is explicitly designed as a weighted moving average unit, which is more effective for video-based applications.

iii) By comparing our DA-RNN with the FCN, DA-RNN achieves better labeling accuracy, thanks to its ability in capturing the temporal information across video frames.

\subsection{Analysis on Network Inputs}

\begin{table*} \setlength{\tabcolsep}{4pt}
	{
		\centering{
			\begin{tabular}{|l||c|c|c|c||c|c|c|c|}
				\hline Methods  & FCN RGB  & FCN Depth & FCN Normal & FCN RGB-D & DA-RNN RGB & DA-RNN Depth & DA-RNN Normal & DA-RNN RGB-D \\
				\hline
				\hline Background & 96.1 & 97.0 & 95.4 & 97.8 & 97.6 & 98.4  & 98.4  & \textbf{98.7} \\
				\hline Bowl       & \textbf{97.0} & 94.8 & 86.5 & 89.8 & 92.7 & 89.8 & 91.8  & 93.1 \\
				\hline Cap        & 79.0 & 86.7 & 86.7 & 82.7 & 84.4  & 88.9  & \textbf{90.5}   & 87.0 \\
				\hline Cereal Box & 87.5 & 88.1 & 58.3 & 88.5 & 88.3  & 90.6 & 90.3  & \textbf{94.2} \\
				\hline Coffee Mug & 75.7 & 81.9 & 83.1 & 82.2 & 86.3  & 83.1  & 86.3  & \textbf{89.4} \\
				\hline Coffee Table & 95.2 & 87.2 & 83.7  & 96.3 & 97.3  &  91.8 & 91.7  & \textbf{97.8} \\
				\hline Office Chair & 71.6 & 79.0 & 74.8  & 82.4 & 77.0  &  84.2 & 84.5  & \textbf{87.5} \\
				\hline Soda Can   & 82.9 & 84.4 & 85.7  & 86.1 & 88.7  & 89.9  & 88.1  & \textbf{90.7} \\
				\hline Sofa       & 92.9 & 94.2 & 92.6   & 96.1 & 95.6  & 95.6 & 96.1  & \textbf{97.9} \\
				\hline Table      & 89.8 & 69.6 & 68.8   & 92.7 & 92.8 & 81.0  & 81.1  & \textbf{94.5} \\
				\hline
				\hline MEAN       & 85.8 & 85.3 &  81.5  & 89.4 & 90.1  & 89.3  & 89.9  & \textbf{93.1} \\
				\hline \multicolumn{5}{|c||}{Improvement of DA-RNN over FCN} & +4.3 & +4.0 & +8.4 & +3.7 \\
				\hline
			\end{tabular}
			\caption{Comparison in network inputs for image pixel labeling on the RGB-D Scene dataset.}
			\label{table:rgbd}
		}
		\vspace{-4mm}
	}
\end{table*}

\begin{table*} \setlength{\tabcolsep}{4pt}
	{
		\centering{
			\begin{tabular}{|l||c|c|c|c||c|c|c|c|}
				\hline Methods  & FCN RGB  & FCN Depth & FCN Normal & FCN RGB-D & DA-RNN RGB & DA-RNN Depth & DA-RNN Normal & DA-RNN RGB-D \\
				\hline
				\hline Background & 99.1 & 99.0 & 98.1 & 99.4 & \textbf{99.5} & 99.3  & 98.8  & \textbf{99.5} \\
				\hline Bottle     & 79.8 & 80.8 & 76.8 & 81.3 & 84.8 & \textbf{86.1} & 83.1  & 84.5 \\
				\hline Can        & 64.5 & 83.7 & 53.2 & 67.1 & 65.2 & \textbf{84.6}  & 81.1 & 66.9 \\
				\hline Cap        & 81.3 & 85.3 & 87.4 & 83.1 & 84.6 & 87.9 & \textbf{91.1} & 83.6 \\
				\hline Keyboard   & 90.2 & 88.9 & 91.3 & 91.3 & 91.2 & 90.6 & 91.6 & \textbf{92.4} \\
				\hline Monitor    & 87.7 & 90.7 & 90.8 & 92.2 & 91.2 & 92.8 & \textbf{93.9} & 93.2 \\
				\hline Mug        & 68.9 & 84.9 & 66.4 & 70.7 & 70.5 & \textbf{85.0} & 81.2 & 70.6 \\
				\hline Table      & 94.9 & 93.7 & 91.5 & 96.0 & 95.8 & 95.1 & 94.2 & \textbf{96.3} \\
				\hline
				\hline MEAN       & 83.3 & 88.4 &  82.0  & 85.1 & 85.3  & \textbf{90.2}  & 89.4  & 85.9 \\
				\hline \multicolumn{5}{|c||}{Improvement of DA-RNN over FCN} & +2.0 & +1.8 & +7.4 & +0.8 \\
				\hline
			\end{tabular}
			\caption{Comparison in network inputs for image pixel labeling on the ShapeNet Scene dataset.}
			\label{table:shapenet}
		}
		\vspace{-6mm}
	}
\end{table*}

We conduct experiments to analyze the effect of different types of network inputs on semantic labeling. These inputs are RGB image, depth image, normal image and RGB-D image. For the depth image, we normalize the depth values between 0 and 255 and copy it three times to feed it into the network whose input has 3 channels. For the normal image, we compute the surface normals from the  depth image, and then convert the surface normal coordinates into a 3-channel image. A pair of the RGB image and the depth image is referred as a RGB-D image. Our single stream network (Fig. \ref{fig:single_stream}) is used to process the RGB image, the depth image or the normal image, while the double stream network (Fig. \ref{fig:double_stream}) is used to process the RGB-D image. Both networks can be turned into a DA-RNN by adding a recurrent layer as in Fig. \ref{fig:rnn}.

Table \ref{table:rgbd} presents the results of our FCN and DA-RNN with different inputs on the RGB-D Scene dataset. i) Using RGB image achieves better performance than using depth image or using normal image. Since the RGB-D Scene dataset consists of a few number of object instances such as two specific bowls or cereal boxes, color is more discriminative than depth and normal. ii) By using RGB-D images with our double stream FCN, the labeling accuracy is improved over using RGB image, depth image or normal image only. iii) Our DA-RNN consistently improves over its FCN counterpart, which demonstrates the advantages of DA-RNNs on the semantic video labeling task.

\begin{figure*}
	\centering
	\includegraphics[height = 0.62\linewidth, width = 0.9\linewidth]{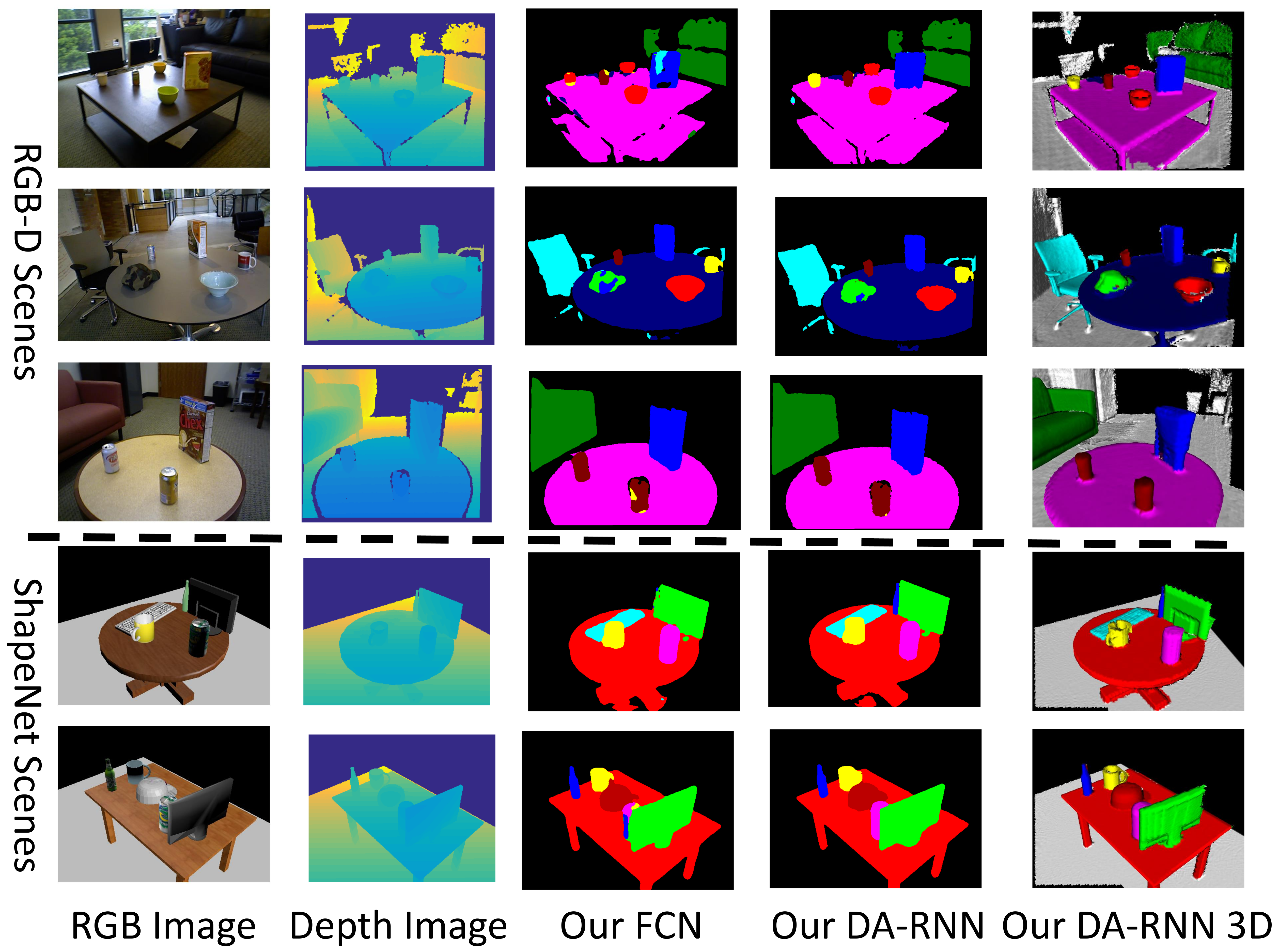}
	\caption{Semantic labeling examples on the RGBS Scene dataset and the ShapeNet Scene dataset.}
	\label{fig:results}
	\vspace{-2mm}
\end{figure*}

Table \ref{table:shapenet} presents the labeling results on the ShapeNet Scene dataset. i) Depth is more discriminative than color. This is because objects in the ShapeNet Scene dataset are sampled from hundreds of 3D shapes with different colors. The objects in the test set are unseen in the training set, so their general shape from the depth images are more discriminative than their color. ii) Combing RGB image and depth image does not improve over using depth image only. It seems that adding color information confuses the network from differentiating objects in different categories but with similar color. iii) Our DA-RNNs achieve better performance than the FCNs consistently across different input types. Fig. \ref{fig:results} shows some labeling examples from our FCN and DA-RNN on the RGB-D Scene dataset and the ShapeNet Scene dataset.

\begin{figure*}
	\centering
	\includegraphics[height = 0.16\linewidth, width = 0.88\linewidth]{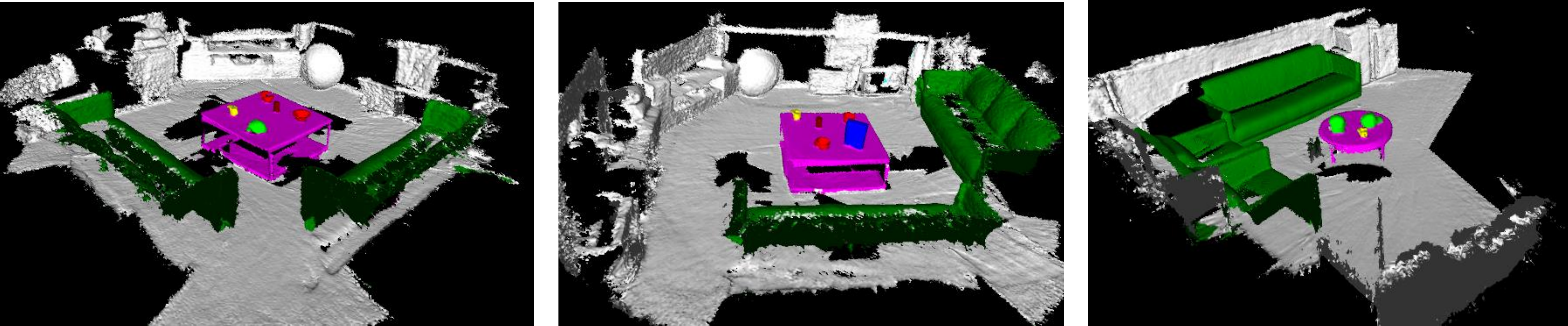}
	\caption{The semantic 3D mapping built by our method using the RGB-D Scene dataset.}
	\label{fig:scenes}
	\vspace{-6mm}
\end{figure*}

\begin{table} \setlength{\tabcolsep}{0.5pt}
	{
		\centering{
			\begin{tabular}{|l||c|c|c|c|}
				\hline Methods  & HMP2D \cite{lai2014unsupervised}  & HMP3D \cite{lai2014unsupervised} & HMP2D+3D \cite{lai2014unsupervised} & DA-RNN RGB-D \\
				\hline
				\hline Background & 42.9 / \textbf{99.6} & \textbf{99.9} / 80.0 & 95.8 / 95.0 & 94.7 / 96.4 \\
				\hline Bowl       & 74.4 / 85.0 & \textbf{100.0} / \textbf{96.2} & 97.0 / 89.1  & 95.3 / 91.0 \\
				\hline Cap        & 74.9 / 98.6 & 91.3 / 98.2 & 82.7 / \textbf{99.0} & \textbf{93.5} / 91.1 \\
				\hline Cereal Box & 79.9 / 98.6 & 85.1 / \textbf{100.0} & 96.2 / 99.3 & \textbf{98.0} / 93.5 \\
				\hline Coffee Mug & 64.4 / 87.8 & 90.0 / \textbf{93.9} & 81.0 / 92.6 & \textbf{90.5} / 86.3 \\
				\hline Coffee Table & 11.9 / 17.9 & 96.1 / \textbf{100.0} & \textbf{98.7} / 98.0  & 93.8 / 97.3 \\
				\hline Office Chair & 17.7 / 17.2 & 57.6 / \textbf{100.0} & 89.7 / 94.5  & \textbf{96.0} / 96.2 \\
				\hline Soda Can   & 78.2 / \textbf{98.1} & \textbf{100.0} / 81.9 &  97.7 / 98.0  & 92.0 / 83.4 \\
				\hline Sofa       & 29.3 / 39.8 & 82.7 / \textbf{100.0} & 92.5 / 92.0   & \textbf{99.6} / 91.3 \\
				\hline Table      & 16.4 / 23.3 & 95.3 / \textbf{98.7} & 97.6 / 96.0   & \textbf{98.2} / 97.2 \\
				\hline
				\hline MEAN       & 49.0 / 66.6 & 89.8 / 94.9 &  92.8 / \textbf{95.3}  & \textbf{95.2} / 92.4 \\
				\hline
			\end{tabular}
			\caption{Precision and recall of 3D point labeling on the RGB-D Scene dataset.}
			\label{table:points}
		}
	}
	\vspace{-6mm}
\end{table}

\subsection{3D Scene Labeling Results}

In this experiment, we evaluate our framework on the 3D point labeling task. Our DA-RNN generates pixel-wise labeling of each video frame that is integrated into the KinectFusion map to label the 3D scene. Since the 3D points provided in the RGB-D Scene dataset are not in the same 3D space of our KinectFusion map, we use the following procedure to obtain the labels of these 3D points. For each 3D point, we project it to all the video frames with camera poses provided by the dataset and then check its visibility in the frames. A 3D point is visible in a frame if its projection is inside the frame and the value of the projection in the depth image is within a range of the depth of the 3D point. Then we accumulate all the visible labels of the 3D point, and use the one with the maximum frequency as the final label for it.

Table \ref{table:points} presents the 3D point labeling precision and recall on the RGB-D Scene dataset, where we compare our method with three variations of the method proposed in \cite{lai2014unsupervised}. It is worth to mention that the models in \cite{lai2014unsupervised} are trained with synthetic data generated from rendering 3D shapes only and tested on all the 14 videos in the RGB-D Scene dataset. To test our method on the same videos, we conduct a two-fold cross validation and obtain the results on all the 14 videos. From Table \ref{table:points}, we can see that our method achieves comparable 3D point labeling precision and recall with the HMP2D+3D model. While \cite{lai2014unsupervised} employs several heuristics to remove the ground plane and the table-top, our system processes a RGB-D video automatically without such heuristics. Fig. \ref{fig:scenes} shows some semantic mapping results on the RGB-D Scene dataset. Please see our project website for the result video on the two datasets. The labeling errors are more often caused by confusion between classes with similar 3D shape such as mug and can. Data association accuracy also affects the performance. For example, we sometimes see that the bottom of an object is labeled as table due to wrong assocation to table pixels in the previous frame.

\section{Conclusion} 

In this work, we introduce DA-RNNs, a novel framework for joint 3D mapping and semantic labeling on RGB-D videos. DA-RNNs integrate a recurrent neural network for video semantic labeling with KinectFusion. To achieve a compact network representation, recurrent reasoning is only performed over the currently visible part of the environment, using data association to define the connectivity between recurrent units. The labels predicted by the RNN are fused into the KinectFusion map for dense semantic mapping. Experiments are conducted on a real world dataset and a synthetic dataset of RGB-D videos. The experimental results and analyses demonstrate the advantages of our method on video semantic labeling and 3D scene mapping.

A key advantage of DA-RNNs is their flexibility. While this paper focuses on object class labeling, we believe that the same architecture could be applied to train networks for a wide range of semantic labeling problems, including object instance and pose detection, material recognition, and physical support estimation. Data association between frames can also be obtained in different ways such as using optical flow methods. Another promising avenue for improvement is the incorporation of shape information provided by the 3D map.

\vspace{-1mm}
\section*{Acknowledgments}

This work was funded in part by ONR grant N00014-13-1-0720 and by Northrop Grumman. We thank Tanner Schmidt for fruitful discussions and for providing his implementation of KinectFusion.

\bibliographystyle{plainnat}
\bibliography{references,fox}

\begin{thebibliography}{36}
\providecommand{\natexlab}[1]{#1}
\providecommand{\url}[1]{\texttt{#1}}
\expandafter\ifx\csname urlstyle\endcsname\relax
  \providecommand{\doi}[1]{doi: #1}\else
  \providecommand{\doi}{doi: \begingroup \urlstyle{rm}\Url}\fi

\bibitem[Abadi et~al.(2016)Abadi, Agarwal, Barham, Brevdo, Chen, Citro,
  Corrado, Davis, Dean, Devin, et~al.]{abadi2016tensorflow}
Mart{\'\i}n Abadi, Ashish Agarwal, Paul Barham, Eugene Brevdo, Zhifeng Chen,
  Craig Citro, Greg~S Corrado, Andy Davis, Jeffrey Dean, Matthieu Devin, et~al.
\newblock Tensorflow: Large-scale machine learning on heterogeneous distributed
  systems.
\newblock \emph{arXiv preprint arXiv:1603.04467}, 2016.

\bibitem[Badrinarayanan et~al.(2015)Badrinarayanan, Kendall, and
  Cipolla]{badrinarayanan2015segnet}
Vijay Badrinarayanan, Alex Kendall, and Roberto Cipolla.
\newblock Segnet: A deep convolutional encoder-decoder architecture for image
  segmentation.
\newblock \emph{arXiv preprint arXiv:1511.00561}, 2015.

\bibitem[Brachmann et~al.(2014)Brachmann, Krull, Michel, Gumhold, Shotton, and
  Rother]{brachmann2014learning}
Eric Brachmann, Alexander Krull, Frank Michel, Stefan Gumhold, Jamie Shotton,
  and Carsten Rother.
\newblock Learning 6{D} object pose estimation using 3{D} object coordinates.
\newblock In \emph{European Conference on Computer Vision (ECCV)}, pages
  536--551, 2014.

\bibitem[Chang et~al.(2015)Chang, Funkhouser, Guibas, Hanrahan, Huang, Li,
  Savarese, Savva, Song, Su, et~al.]{chang2015shapenet}
Angel~X Chang, Thomas Funkhouser, Leonidas Guibas, Pat Hanrahan, Qixing Huang,
  Zimo Li, Silvio Savarese, Manolis Savva, Shuran Song, Hao Su, et~al.
\newblock Shapenet: An information-rich 3{D} model repository.
\newblock \emph{arXiv preprint arXiv:1512.03012}, 2015.

\bibitem[Chen et~al.(2016)Chen, Papandreou, Kokkinos, Murphy, and
  Yuille]{chen2016deeplab}
Liang-Chieh Chen, George Papandreou, Iasonas Kokkinos, Kevin Murphy, and Alan~L
  Yuille.
\newblock Deeplab: Semantic image segmentation with deep convolutional nets,
  atrous convolution, and fully connected {CRF}s.
\newblock \emph{arXiv preprint arXiv:1606.00915}, 2016.

\bibitem[Cho et~al.(2014)Cho, Van~Merri{\"e}nboer, Bahdanau, and
  Bengio]{cho2014properties}
Kyunghyun Cho, Bart Van~Merri{\"e}nboer, Dzmitry Bahdanau, and Yoshua Bengio.
\newblock On the properties of neural machine translation: Encoder-decoder
  approaches.
\newblock \emph{arXiv preprint arXiv:1409.1259}, 2014.

\bibitem[Crandall et~al.(2011)Crandall, Owens, Snavely, and
  Huttenlocher]{crandall2011discrete}
David Crandall, Andrew Owens, Noah Snavely, and Dan Huttenlocher.
\newblock Discrete-continuous optimization for large-scale structure from
  motion.
\newblock In \emph{IEEE Conference on Computer Vision and Pattern Recognition
  (CVPR)}, pages 3001--3008, 2011.

\bibitem[Deng et~al.(2009)Deng, Dong, Socher, Li, Li, and
  Fei-Fei]{deng2009imagenet}
Jia Deng, Wei Dong, Richard Socher, Li-Jia Li, Kai Li, and Li~Fei-Fei.
\newblock Imagenet: A large-scale hierarchical image database.
\newblock In \emph{IEEE Conference on Computer Vision and Pattern Recognition
  (CVPR)}, pages 248--255, 2009.

\bibitem[Felzenszwalb et~al.(2008)Felzenszwalb, McAllester, and
  Ramanan]{felzenszwalb2008discriminatively}
Pedro Felzenszwalb, David McAllester, and Deva Ramanan.
\newblock A discriminatively trained, multiscale, deformable part model.
\newblock In \emph{IEEE Conference on Computer Vision and Pattern Recognition
  (CVPR)}, pages 1--8, 2008.

\bibitem[Girshick et~al.(2014)Girshick, Donahue, Darrell, and
  Malik]{girshick2014rich}
Ross Girshick, Jeff Donahue, Trevor Darrell, and Jitendra Malik.
\newblock Rich feature hierarchies for accurate object detection and semantic
  segmentation.
\newblock In \emph{IEEE Conference on Computer Vision and Pattern Recognition
  (CVPR)}, pages 580--587, 2014.

\bibitem[Graves et~al.(2013)Graves, Jaitly, and Mohamed]{graves2013hybrid}
Alex Graves, Navdeep Jaitly, and Abdel-rahman Mohamed.
\newblock Hybrid speech recognition with deep bidirectional {LSTM}.
\newblock In \emph{IEEE Workshop on Automatic Speech Recognition and
  Understanding (ASRU)}, pages 273--278, 2013.

\bibitem[Henry et~al.(2012)Henry, Krainin, Herbst, Ren, and Fox]{henry2012rgb}
Peter Henry, Michael Krainin, Evan Herbst, Xiaofeng Ren, and Dieter Fox.
\newblock {RGB-D} mapping: Using {K}inect-style depth cameras for dense 3{D}
  modeling of indoor environments.
\newblock \emph{The International Journal of Robotics Research}, 31\penalty0
  (5):\penalty0 647--663, 2012.

\bibitem[Henry et~al.(2013)Henry, Fox, Bhowmik, and Mongia]{henry2013patch}
Peter Henry, Dieter Fox, Achintya Bhowmik, and Rajiv Mongia.
\newblock Patch volumes: Segmentation-based consistent mapping with {RGB-D}
  cameras.
\newblock In \emph{International Conference on 3D Vision (3DV)}, pages
  398--405, 2013.

\bibitem[Hochreiter and Schmidhuber(1997)]{hochreiter1997long}
Sepp Hochreiter and J{\"u}rgen Schmidhuber.
\newblock Long short-term memory.
\newblock \emph{Neural computation}, 9\penalty0 (8):\penalty0 1735--1780, 1997.

\bibitem[Keller et~al.(2013)Keller, Lefloch, Lambers, Izadi, Weyrich, and
  Kolb]{keller2013real}
Maik Keller, Damien Lefloch, Martin Lambers, Shahram Izadi, Tim Weyrich, and
  Andreas Kolb.
\newblock Real-time 3{D} reconstruction in dynamic scenes using point-based
  fusion.
\newblock In \emph{International Conference on 3D Vision (3DV)}, pages 1--8,
  2013.

\bibitem[Kr{\"a}henb{\"u}hl and Koltun(2011)]{krahenbuhl2011efficient}
Philipp Kr{\"a}henb{\"u}hl and Vladlen Koltun.
\newblock Efficient inference in fully connected {CRF}s with gaussian edge
  potentials.
\newblock In \emph{Advances in Neural Information Processing Systems (NIPS)},
  pages 109--117, 2011.

\bibitem[Lai(2013)]{Lai13Obj}
Kevin Lai.
\newblock \emph{Object Recognition and Semantic Scene Labeling for {RGB-D}
  Data}.
\newblock PhD thesis, University of Washington, 12 2013.

\bibitem[Lai et~al.(2014)Lai, Bo, and Fox]{lai2014unsupervised}
Kevin Lai, Liefeng Bo, and Dieter Fox.
\newblock Unsupervised feature learning for 3{D} scene labeling.
\newblock In \emph{IEEE International Conference on Robotics and Automation
  (ICRA)}, pages 3050--3057, 2014.

\bibitem[Long et~al.(2015)Long, Shelhamer, and Darrell]{long2015fully}
Jonathan Long, Evan Shelhamer, and Trevor Darrell.
\newblock Fully convolutional networks for semantic segmentation.
\newblock In \emph{IEEE Conference on Computer Vision and Pattern Recognition
  (CVPR)}, pages 3431--3440, 2015.

\bibitem[McCormac et~al.(2016)McCormac, Handa, Davison, and
  Leutenegger]{mccormac2016semanticfusion}
John McCormac, Ankur Handa, Andrew Davison, and Stefan Leutenegger.
\newblock Semantic{F}usion: Dense 3{D} semantic mapping with convolutional
  neural networks.
\newblock \emph{arXiv preprint arXiv:1609.05130}, 2016.

\bibitem[Newcombe et~al.(2011)Newcombe, Izadi, Hilliges, Molyneaux, Kim,
  Davison, Kohi, Shotton, Hodges, and Fitzgibbon]{newcombe2011kinectfusion}
Richard~A Newcombe, Shahram Izadi, Otmar Hilliges, David Molyneaux, David Kim,
  Andrew~J Davison, Pushmeet Kohi, Jamie Shotton, Steve Hodges, and Andrew
  Fitzgibbon.
\newblock Kinect{F}usion: Real-time dense surface mapping and tracking.
\newblock In \emph{International Symposium on Mixed and Augmented Reality
  (ISMAR)}, pages 127--136, 2011.

\bibitem[Pavel et~al.(2015)Pavel, Schulz, and Behnke]{pavel2015recurrent}
Mircea~Serban Pavel, Hannes Schulz, and Sven Behnke.
\newblock Recurrent convolutional neural networks for object-class segmentation
  of {RGB-D} video.
\newblock In \emph{International Joint Conference on Neural Networks (IJCNN)},
  pages 1--8, 2015.

\bibitem[Ren et~al.(2012)Ren, Bo, and Fox]{ren2012rgb}
Xiaofeng Ren, Liefeng Bo, and Dieter Fox.
\newblock {RGB-D} scene labeling: Features and algorithms.
\newblock In \emph{IEEE Conference on Computer Vision and Pattern Recognition
  (CVPR)}, pages 2759--2766, 2012.

\bibitem[Salas-Moreno et~al.(2013)Salas-Moreno, Newcombe, Strasdat, Kelly, and
  Davison]{salas2013slam++}
Renato Salas-Moreno, Richard Newcombe, Hauke Strasdat, Paul Kelly, and Andrew
  Davison.
\newblock {SLAM}++: Simultaneous localisation and mapping at the level of
  objects.
\newblock In \emph{IEEE Conference on Computer Vision and Pattern Recognition
  (CVPR)}, pages 1352--1359, 2013.

\bibitem[Savarese and Fei-Fei(2007)]{savarese20073d}
Silvio Savarese and Li~Fei-Fei.
\newblock 3{D} generic object categorization, localization and pose estimation.
\newblock In \emph{International Conference on Computer Vision (ICCV)}, pages
  1--8, 2007.

\bibitem[Shelhamer et~al.(2016)Shelhamer, Rakelly, Hoffman, and
  Darrell]{shelhamer2016clockwork}
Evan Shelhamer, Kate Rakelly, Judy Hoffman, and Trevor Darrell.
\newblock Clockwork convnets for video semantic segmentation.
\newblock In \emph{Video Semantic Segmentation Workshop at European Conference
  in Computer Vision (ECCV)}, pages 852--868, 2016.

\bibitem[Shotton et~al.(2006)Shotton, Winn, Rother, and
  Criminisi]{shotton2006textonboost}
Jamie Shotton, John Winn, Carsten Rother, and Antonio Criminisi.
\newblock Textonboost: Joint appearance, shape and context modeling for
  multi-class object recognition and segmentation.
\newblock In \emph{European Conference on Computer Vision (ECCV)}, pages 1--15,
  2006.

\bibitem[Simonyan and Zisserman(2014)]{simonyan2014very}
Karen Simonyan and Andrew Zisserman.
\newblock Very deep convolutional networks for large-scale image recognition.
\newblock \emph{arXiv preprint arXiv:1409.1556}, 2014.

\bibitem[Snavely et~al.(2008)Snavely, Seitz, and Szeliski]{snavely2008skeletal}
Noah Snavely, Steven~M Seitz, and Richard Szeliski.
\newblock Skeletal graphs for efficient structure from motion.
\newblock In \emph{IEEE Conference on Computer Vision and Pattern Recognition
  (CVPR)}, 2008.

\bibitem[Song and Xiao(2016)]{song2016deep}
Shuran Song and Jianxiong Xiao.
\newblock Deep sliding shapes for amodal 3{D} object detection in {RGB-D}
  images.
\newblock In \emph{IEEE Conference on Computer Vision and Pattern Recognition
  (CVPR)}, pages 808--816, 2016.

\bibitem[Sundermeyer et~al.(2012)Sundermeyer, Schl{\"u}ter, and
  Ney]{sundermeyer2012lstm}
Martin Sundermeyer, Ralf Schl{\"u}ter, and Hermann Ney.
\newblock {LSTM} neural networks for language modeling.
\newblock In \emph{Interspeech}, pages 194--197, 2012.

\bibitem[Sutskever et~al.(2014)Sutskever, Vinyals, and
  Le]{sutskever2014sequence}
Ilya Sutskever, Oriol Vinyals, and Quoc~V Le.
\newblock Sequence to sequence learning with neural networks.
\newblock In \emph{Advances in Neural Information Processing Systems (NIPS)},
  pages 3104--3112, 2014.

\bibitem[Whelan et~al.(2012)Whelan, Kaess, Fallon, Johannsson, Leonard, and
  McDonald]{Whe12Kin}
Thomas Whelan, Michael Kaess, Maurice Fallon, Hordur Johannsson, John Leonard,
  and John McDonald.
\newblock {Kintinuous: Spatially Extended KinectFusion}.
\newblock In \emph{RSS Workshop on RGB-D: Advanced Reasoning with Depth
  Cameras}, 2012.

\bibitem[Whelan et~al.(2015)Whelan, Leutenegger, Salas-Moreno, Glocker, and
  Davison]{whelan2015elasticfusion}
Thomas Whelan, Stefan Leutenegger, Renato~F Salas-Moreno, Ben Glocker, and
  Andrew~J Davison.
\newblock Elastic{F}usion: Dense {SLAM} without a pose graph.
\newblock In \emph{Robotics: science and systems}, volume~11, 2015.

\bibitem[Xiang and Savarese(2012)]{xiang2012estimating}
Yu~Xiang and Silvio Savarese.
\newblock Estimating the aspect layout of object categories.
\newblock In \emph{IEEE Conference on Computer Vision and Pattern Recognition
  (CVPR)}, pages 3410--3417, 2012.

\bibitem[Zheng et~al.(2015)Zheng, Jayasumana, Romera-Paredes, Vineet, Su, Du,
  Huang, and Torr]{zheng2015conditional}
Shuai Zheng, Sadeep Jayasumana, Bernardino Romera-Paredes, Vibhav Vineet,
  Zhizhong Su, Dalong Du, Chang Huang, and Philip~HS Torr.
\newblock Conditional random fields as recurrent neural networks.
\newblock In \emph{IEEE International Conference on Computer Vision (ICCV)},
  pages 1529--1537, 2015.

\end{thebibliography}
	
\end{document}